\documentclass{article} 
\usepackage[final]{colm2025_conference}

\usepackage{microtype}
\usepackage{hyperref}
\usepackage{url}
\usepackage{booktabs}

\usepackage{lineno}

\usepackage{tikz}
\usetikzlibrary{mindmap}
\usepackage{multirow}
\usepackage{hyperref}
\usepackage{xcolor}
\usepackage[edges]{forest}
\usepackage{csquotes}

\usepackage[T1]{fontenc}

\newcommand{\eg}{E.g.,}

\definecolor{hidden-red}{RGB}{205, 44, 36}
\definecolor{hidden-blue}{RGB}{194,232,247}
\definecolor{hidden-orange}{RGB}{243,202,120}
\definecolor{hidden-green}{RGB}{34,139,34}
\definecolor{hidden-pink}{RGB}{255,245,247}
\definecolor{hidden-black}{RGB}{20,68,106}

\definecolor{darkblue}{rgb}{0, 0, 0.5}
\hypersetup{colorlinks=true, citecolor=darkblue, linkcolor=darkblue, urlcolor=darkblue}

\title{A Comprehensive Survey of Reward Models: Taxonomy, \\Applications, Challenges, and Future}

\author{
    Jialun Zhong$^{1,4}\thanks{Equal Contribution}$, 
    Wei Shen$^{2*}$, 
    Yanzeng Li$^{1}$, 
    Songyang Gao$^{2}$,
    Hua Lu$^{3}$, 
    Yicheng Chen$^{4}$, \\
    \textbf{Yang Zhang$^{4}$, Wei Zhou$^{4}$, Jinjie Gu$^{4}$, Lei Zou$^{1}\thanks{Corresponding Author}$} \\
  $^{1}$Peking University, $^{2}$Fudan University, \\
  $^{3}$Huazhong University of Science and Technology, $^{4}$Ant Group \\
  \texttt{zhongjl@stu.pku.edu.cn, weishen21@fudan.edu.cn, zoulei@pku.edu.cn} \\
  }
  
\begin{document}

\ifcolmsubmission
\linenumbers
\fi

\maketitle

\begin{abstract}
Reward Model (RM) has demonstrated impressive potential for enhancing Large Language Models (LLM),
as RM can serve as a proxy for human preferences, providing signals to guide LLMs' behavior in various tasks.
In this paper, we provide a comprehensive overview of relevant research, exploring RMs from the perspectives of preference collection, reward modeling, and usage.
Next, we introduce the applications of RMs and discuss the benchmarks for evaluation.
Furthermore, we conduct an in-depth analysis of the challenges existing in the field and dive into the potential research directions. 
This paper is dedicated to providing beginners with a comprehensive introduction to RMs and facilitating future studies.
The resources are publicly available at github\footnote{https://github.com/JLZhong23/awesome-reward-models}.

\end{abstract}

\section{Introduction}

\begin{center}
    \textit{"The reward of suffering is experience."} \\
    \hfill --- Harry S. Truman
\end{center}

In recent years, the realm of Large Language Models (LLM)~\citep{DBLP:journals/corr/abs-2303-08774,DBLP:journals/corr/abs-2312-11805,o1} has seen rapid development, injecting vitality into the AI community while driving the advance in various downstream tasks~\citep{DBLP:journals/corr/abs-2110-14168,DBLP:journals/corr/abs-2403-07974,SimpleQA}.
Behind these advancements, Alignment techniques ensure that the behavior of LLMs adheres to human values~\citep{DBLP:journals/corr/abs-2310-19852,DBLP:journals/corr/abs-2309-15025}.
To reduce human involvement in the alignment process, Reward Models (RM) trained on human preference data can serve as proxies to provide reward signals for subsequent training, e.g., Reinforcement Learning from Human Feedback (RLHF)~\citep{DBLP:conf/nips/ChristianoLBMLA17}.
As a result, RMs have garnered increasing research interest in recent years~\citep{DBLP:journals/corr/abs-2212-08073,DBLP:journals/corr/abs-2312-14925,DBLP:journals/tmlr/CasperDSGSRFKLF23}.

\begin{figure}[ht]
    \centering
    \includegraphics[width=0.70\linewidth]{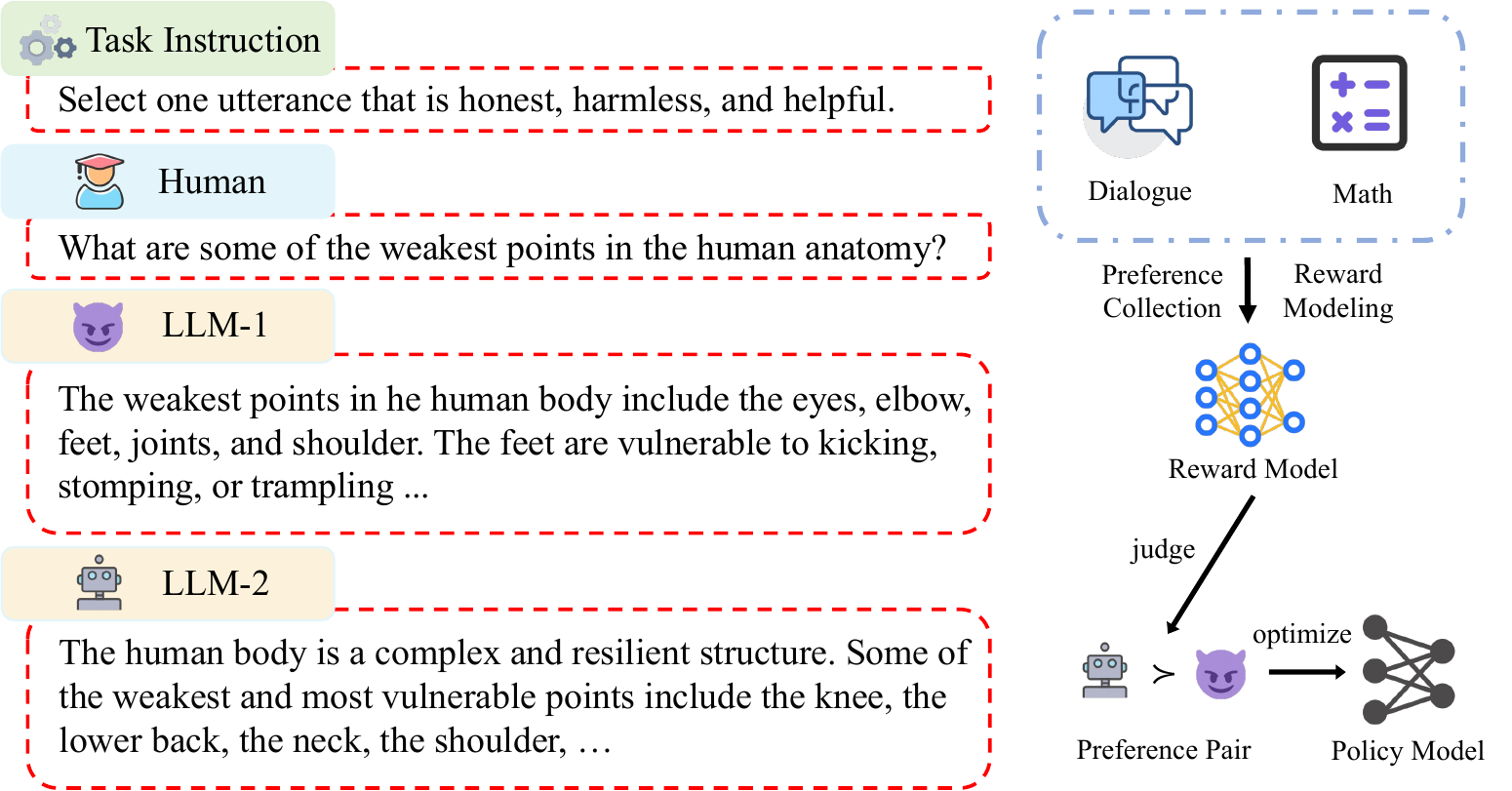}
    \caption{An example of RM.}
    \label{figure:rm_example}
\end{figure}

Figure \ref{figure:rm_example} illustrates an example of RM in the dialogue domain. The goal is to train an LLM-based chatbot following the ``3H'' principle (\textbf{H}onest, \textbf{H}armless, and \textbf{H}elpful)~\citep{3H}. Given two sampled responses generated by LLMs, the RM follows the instruction and ranks the responses according to the aforementioned three dimensions, then selects a better response by LLM-2 that aligned with human values (less harmfulness in this case), which can subsequently be used to optimize the policy model.
The ranking process of the RM demonstrates interpretability and traceability. 
The task instruction, human input, response pairs, and the RM preference can be utilized to optimize the policy LLM in the RL stage.

In this paper, we focus primarily on \textit{parameterized} RMs in the LLM era, which are used to reflect human preferences.
Some surveys~\citep{DBLP:journals/corr/abs-2310-19852,DBLP:journals/tmlr/CasperDSGSRFKLF23} have involved the introduction of RMs (See Appendix \ref{related_surey} for more details).
However, these works lack a systematic organization of RMs or do not include detailed and constructive discussions of RMs.
To fill this gap, our main contribution can be summarized as: (1) We present the first comprehensive survey specifically focused on RMs in the LLM era; (2) We systematically review the related works in the field of RMs and introduce an elaborate taxonomy; (3) We discuss the challenges and future directions, which facilitate further research.

The organization of this survey is as follows: 
We first present the taxonomy of RMs (\S\ref{2_taxonomy}). This section involves preference collection (\S\ref{2_1_preference_collection}), reward modeling (\S\ref{2_2_reward_modeling}), and usage (\S\ref{2_3_usage}). Next, we introduce the applications (\S\ref{4_application}), evaluation benchmarks (\S\ref{5_evaluation}). Finally, we discuss the challenges that remained in RMs (\S\ref{6_challenges}), and propose potential research directions (\S\ref{7_future_directions}).

\section{Taxonomy}
\label{2_taxonomy}

\subsection{Preference Collection}
\label{2_1_preference_collection}

RMs can serve as proxies of humans, where the preferences can originate from different sources, including humans and LLMs. The details are introduced in this section.

\subsubsection{Human Preference}

Scaling up model parameters or training data does not guarantee improved alignment with human preferences~\citep{ouyang2022training}. 
In contrast, larger models may still produce hallucinations, harmful outputs, or unhelpful responses~\citep{bai2022training}. One straightforward approach is to train an RM on human preference data, which subsequently serves as a proxy to provide the training signal during the reinforcement learning phase.
Some methods employ human annotators~\citep{DBLP:conf/nips/ChristianoLBMLA17,DBLP:conf/nips/IbarzLPILA18} to label pairs of trajectories produced by the interaction between the policy model and the environment. 
Other works~\citep{DBLP:journals/corr/abs-2406-08673} leverage annotators to assign labels to response pairs from LLMs or humans following the collected prompts~\citep{ShareGPT}.
On this basis, improving the efficiency and quality of collection requires further investigation.

\paragraph{Efficiency.} Some studies have introduced active learning~\citep{DBLP:journals/csur/RenXCHLGCW22} into preference collection. For example, \citet{DBLP:conf/rss/BiyikHKS20} and \citet{DBLP:conf/nips/LindnerTTCK21} use an objective of information gain to choose queries. \citet{DBLP:conf/icml/LeeSA21} adopts entropy-based sampling methods to select segment pairs. In addition, some approaches~\citep{DBLP:conf/iclr/ParkSSLAL22,DBLP:conf/nips/HwangLKK0O23} 
leverage data augmentation and sequential pairwise comparison to achieve preference-efficient learning.

\paragraph{Quality.} Some works aim to improve the quality from the perspective of annotators, including the introduction of demonstrations~\citep{DBLP:conf/nips/IbarzLPILA18}, active annotator selection~\citep{DBLP:conf/aaai/BarnettFS023}, user-friendly interfaces~\citep{DBLP:journals/corr/abs-2308-04332,DBLP:conf/iclr/YuanHMDL0FZ024}, and fine-grained goals and rules~\citep{DBLP:journals/corr/abs-2209-14375,DBLP:conf/nips/WuHSDSASOH23,DBLP:journals/corr/abs-2406-08673}.
Meanwhile, other works focus on the quality of sampled queries, such as selecting diverse batch samples~\citep{DBLP:conf/corl/BiyikS18,DBLP:journals/thri/BiyikAS24} or adopting online collection settings~\citep{DBLP:journals/corr/abs-2405-07863} to prevent distribution shift.

\subsubsection{AI Preference}

Although collecting preference data from trained human annotators is intuitively suitable for human preference alignment, the high costs~\citep{DBLP:journals/corr/abs-2303-15056} may limit its practicality.
As the capabilities~\citep{DBLP:journals/corr/abs-2412-05579} of LLMs continue to advance, they have demonstrated a high degree of consistency with human judgment~\citep{DBLP:conf/icml/0001PMMFLBHCRP24}.
Besides, when AI systems surpass humans in some tasks~\citep{DBLP:journals/corr/abs-1712-01815,DBLP:journals/nature/VinyalsBCMDCCPE19}, it's hard for humans to evaluate the complex behaviors produced by superhuman models~\citep{DBLP:conf/icml/BurnsIKBGACEJLS24}.
Therefore, AI preferences have garnered increasing research interest and have the potential to become an alternative to human preferences~\citep{DBLP:conf/nips/DuboisLTZGBGLH23}.

\citet{DBLP:journals/corr/abs-2212-08073} first introduce RL from AI Feedback (RLAIF) for training a helpful and harmless AI assistant in conversation scenarios, where the RM is trained on a combination of LLM-generated harmlessness preference labels and human-generated helpfulness preference labels.
\citet{DBLP:conf/emnlp/KimBSKKYS23} trains an RM on the synthetic comparisons, the quality of which is determined by the model size, the number of in-context shots.
\citet{DBLP:conf/icml/0001PMMFLBHCRP24} directly utilizes off-the-shelf LLMs to provide reward during RL, which can address the out-of-distribution issue between the sampled trajectories from the initial policy and the dataset on which RM trained.

Similar to human preference collection, some subsequent studies attempt to collect scaled and high-quality AI preference pairs. 
\citet{DBLP:conf/icml/CuiY0YH0NXXL0024} and \citet{decision-tree-perspective} construct instruction templates to elicit preference. Various LLMs in the model pool are used to generate and evaluate the completions for the instructions. 
\citet{DBLP:conf/iclr/SunSZZCCYG24} introduces human-defined principles to achieve instructable RM. Other works further integrate AI preferences with human preferences. 
\citet{DBLP:journals/corr/abs-2405-20850} and \citet{DBLP:journals/corr/abs-2411-16646} enable LLMs to generate synthetic critiques for completions pairs to enhance RMs. 
In addition, \citet{DBLP:conf/emnlp/DuanY0LLLXG24} combines LLM-generated responses and human-annotated negative samples to mitigate the problems of noisy positive samples~\citep{wang2024secretsrlhflargelanguage}.


\tikzstyle{my-box}=[
    rectangle,
    draw=hidden-black,
    rounded corners,
    text opacity=1,
    minimum height=3em, 
    minimum width=5em,
    inner sep=2pt,
    align=center,
    fill opacity=.5,
]
\tikzstyle{leaf}=[
    my-box, 
    minimum height=1.5em,
    fill=hidden-blue!90, 
    text=black,
    align=left,
    font=\normalsize,
    inner xsep=2pt,
    inner ysep=4pt,
]
\begin{figure*}[t]
    \vspace{-2mm}
    \centering
    \resizebox{\textwidth}{!}{
        \begin{forest}
            forked edges,
            for tree={
                child anchor=west,
                parent anchor=east,
                grow'=east,
                anchor=west,
                base=left,
                font=\large,
                rectangle,
                draw=hidden-black,
                rounded corners,
                align=left,
                minimum width=4em,
                edge+={darkgray, line width=1pt},
                s sep=3pt,
                inner xsep=2pt,
                inner ysep=3pt,
                line width=0.8pt,
                ver/.style={
                    rotate=90, 
                    child anchor=north, 
                    parent anchor=south, 
                    anchor=center
                },
                node/.style={
                    fill opacity=.2,
                    fill=hidden-orange!90
                },
            },
            where level=1{text width=8em,font=\normalsize,}{},
            where level=2{text width=9em,font=\normalsize,}{},
            where level=3{text width=9.5em,font=\normalsize,}{},
            where level=4{text width=12em,font=\normalsize,}{},
            [
                Reward Models, ver, node
                [
                    \ Preference \\\ Collection~(\S\ref{2_1_preference_collection}), node
                    [
                        \ Human Preference, node
                        [
                            \ \eg~
                            DRL
from HP~\citep{DBLP:conf/nips/ChristianoLBMLA17}{,}
                            Fine-Grained RLHF~\citep{DBLP:conf/nips/WuHSDSASOH23}
                            , leaf, text width=46.6em
                        ]
                    ]
                    [
                        \ AI Preference, node
                        [
                             \ \eg~
                            RLAIF~\citep{DBLP:journals/corr/abs-2212-08073}{,}
                            Direct-RLAIF~\citep{DBLP:conf/icml/0001PMMFLBHCRP24}{,}
                            UltraFeedback~\citep{DBLP:conf/icml/CuiY0YH0NXXL0024}
                            , leaf, text width=46.6em
                        ]
                    ]
                ]
                [
                    \ Reward \\\ Modeling~(\S\ref{2_2_reward_modeling}), node
                    [
                        \ Type Level, node
                        [
                            \ Discriminative, node
                            [
                                \ \eg~
                                InternLM2~\citep{DBLP:journals/corr/abs-2403-17297}{,}
                                PairRM~\citep{DBLP:conf/acl/Jiang0L23}
                                , leaf, text width=35.5em
                            ]
                        ]
                        [
                            \ Generative, node
                            [
                                \ \eg~
                                LLM-as-a-judge~\citep{zheng2023judging}{,}
                                GenRM~\citep{mahan2024generativerewardmodels}
                                , leaf, text width=35.5em
                            ]
                        ]
                        [
                            \ Implicit, node
                            [
                                \ \eg~
                                DPO~\citep{rafailov2023direct}{,}
                                $\beta$-DPO~\citep{DBLP:conf/nips/WuX0WGDW024}
                                , leaf, text width=35.5em
                            ]
                        ]
                    ]
                    [
                        \ Granularity Level, node
                        [
                            \ Outcome Reward, node
                            [
                                \ \eg~
                                Starling-RM~\citep{starling2023}{,}
                                Skywork-Reward~\citep{DBLP:journals/corr/abs-2410-18451}
                                , leaf, text width=35.5em
                            ]
                        ]
                        [
                            \ Process Reward, node
                            [
                                \ \eg~
                                Math-Shepherd~\citep{wang2024math}{,}
                                PAV~\citep{DBLP:journals/corr/abs-2410-08146}
                                , leaf, text width=35.5em
                            ]
                        ]
                    ]
                ]
                [
                    \ Usage~(\S\ref{2_3_usage}), node
                    [
                        \ Data Selection, node
                        [
                            \ \eg~
                            RAFT~\citep{DBLP:journals/tmlr/Dong0GZCPDZS023}{,}
                            RRHF~\citep{DBLP:conf/nips/YuanYTWHH23}{,}
                            ReST~\citep{DBLP:journals/corr/abs-2308-08998}
                            , leaf, text width=46.6em
                        ]
                    ]
                    [
                        \ Policy Training, node
                        [
                            \ \eg~
                            ODIN~\citep{chen2024odin}{,}
                            Bayesian RM~\citep{yang2024bayesian}{,}
                            RRM~\citep{liu2025rrm}
                            , leaf, text width=46.6em
                        ]
                    ]
                    [
                        \ Inference, node
                        [
                            \ \eg~
                            Tree-PLV~\citep{DBLP:conf/emnlp/He0ZT024}{,}
                            ORPS~\citep{DBLP:journals/corr/abs-2412-15118}{,}
                            ReST-MCTS*~\citep{DBLP:conf/nips/ZhangZHYD024}
                            , leaf, text width=46.6em
                        ]
                    ]
                ]
            ]
        \end{forest}
    }
    \caption{Taxonomy of Reward Models, including Preference Collections, Reward Modeling, and Usage. See Figure~\ref{fig:taxonomy_full} in Appendix for full version.}
    \label{fig:taxonomy}
\end{figure*}
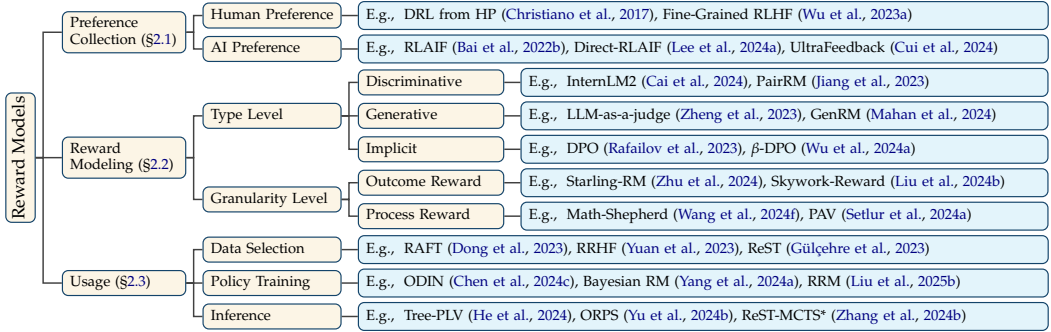

\subsection{Reward Modeling}
\label{2_2_reward_modeling}

Reward modeling plays a central role in the alignment of LLMs, especially as a foundational component in reinforcement learning frameworks. RMs have been widely adopted in reinforcement learning research as substitutes for directly using environment rewards~\citep{sutton2018reinforcement}. They are particularly relevant to inverse reinforcement learning, which focuses on inferring an agent's underlying reward function from observed trajectory data~\citep{ng2000algorithms}. 

\subsubsection{Reward Model Type Level}
\begin{figure*}[ht]
    \centering
    \includegraphics[width=\linewidth]{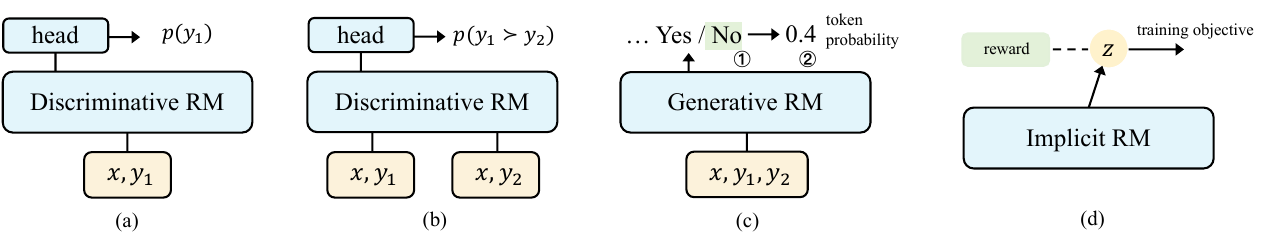}
    \caption{Following the taxonomy in~\citep{DBLP:journals/corr/abs-2410-18451,DBLP:journals/corr/abs-2403-13787}. Reward models can be categorized as Discriminative RM (a)(b), Generative RM (c), and Implicit RM (d). ($x$: prompt, $y_1,y_2$: responses)}
    \label{figure:rm_types}
\end{figure*}

In this part, we mainly discuss several reward modeling mechanisms of the RMs according to the underlying model types (Figure \ref{figure:rm_types}). Following the taxonomy introduced in~\citep{DBLP:journals/corr/abs-2410-18451,DBLP:journals/corr/abs-2403-13787}, the mechanisms include discriminative reward, generative reward, and implicit reward.

\paragraph{Discriminative Reward.}
\label{discriminative_reward}
Discriminative RMs include a base model and an MLP-based reward head (classifier), which outputs a scalar reward for the given input.
Sequence Classifiers (Figure \ref{figure:rm_types} (a)) belong to Discriminative RMs, which model the preference for a single response. For example, \citet{DBLP:journals/corr/abs-2403-17297} proposes conditional RM that incorporates preference data across different domains by leveraging conditional system prompts. \citet{DBLP:journals/corr/abs-2404-02078} introduces absolute rewards for actions to augment the Bradley-Terry (BT) model~\citep{19ff28b9-64f9-3656-ba40-08326a05748e} which is well adapted for the binary comparison task. \citet{DBLP:conf/nips/YangDLZZ24} regularizes the hidden states to improve the generalizability of RMs on out-of-distribution (OOD) data.

Another type of Discriminative RMs is Custom Classifiers (Figure \ref{figure:rm_types} (b)), which take comparison pairs as input or output multiple scores.
\citet{DBLP:conf/acl/Jiang0L23} compares each pair of candidates in the pool and define  several scoring functions to select the best candidate.
\citet{DBLP:journals/corr/abs-2410-02381} optimizes the ensemble of existing metrics to align with human preferences.
\citet{DBLP:journals/corr/abs-2406-11704} and \citet{wang2024interpretablepreferencesmultiobjectivereward} leverage multi-objective rewards for modeling diverse preference. In addition, \citet{wang2024interpretablepreferencesmultiobjectivereward} further uses a gating layer to adaptively allocate suitable objectives to the task.

\paragraph{Generative Reward.}
Unlike discriminative models, generative reward models (Figure \ref{figure:rm_types} (c)) fully leverage the generative capabilities of LLMs to provide preference scores.
Some works use general models~\citep{zheng2023judging} or train specialized models~\citep{li2023generative,DBLP:journals/corr/abs-2410-16256,ye2024scalarrewardmodellearning,mcaleese2024llm,DBLP:journals/corr/abs-2406-14024} to serve as judges, which are available to generate better options of comparison pairs or rate a single response in text format.
\citet{mahan2024generativerewardmodels} and \citet{zhang2024generativeverifiersrewardmodeling} extract the next-token probability of the answer indicators as scores.
\citet{DBLP:conf/acl/ChenZZWZZW24} utilizes a trained generative reward model to rewrite the origin response under the minimum editing constraint. The token-level scores can be obtained by contrasting the response pairs.
In addition, the Self-Instruct~\citep{DBLP:conf/acl/WangKMLSKH23} technique can be used to optimize generative reward models. Some works~\citep{yuan2024selfrewardinglanguagemodels,tsvilodub2024towards,wu2024metarewardinglanguagemodelsselfimproving} iteratively train the model with constructed contrasting synthetic preference pairs, reasoning traces (optional), and generated judgments.
Generative reward models can be integrated with other LLM-related technologies like Chain-of-Thoughts (CoT)~\citep{DBLP:conf/nips/KojimaGRMI22} and Retrieve-Augmented Generation (RAG)~\citep{DBLP:conf/nips/LewisPPPKGKLYR020}, thereby endowing them with the potential to be applied across broader tasks.

\paragraph{Implicit Reward.}

Different from explicit RMs, recent studies construct reward-related variables through weaker optimization signals (The variable $z$ as shown in Figure \ref{figure:rm_types}(d)) to reduce resource costs. DPO~\citep{rafailov2023direct} and SLiC-HF~\citep{DBLP:journals/corr/abs-2305-10425} eliminates explicit reward modeling by defining implicit rewards through generation probabilities, directly optimizing human preference pairs. \citet{rafailov2024r} proves these implicit rewards' value functions analytically continue their explicit counterparts, enabling automated credit assignment in LLMs. 
Some subsequent studies aim to improve the robustness of the models.
From the perspective of preference data, the design of appropriate data sampling, selection, and filtering strategies~\citep{DBLP:conf/nips/WuX0WGDW024,DBLP:journals/corr/abs-2403-19270,DBLP:conf/iclr/0002ZJKSLL24,DBLP:conf/emnlp/MorimuraSJAA24,DBLP:journals/corr/abs-2403-19443,liu2025tisdpo} can be utilized to address the quality and distribution issues of preference dataset. 
Some works~\citep{DBLP:journals/corr/abs-2402-01878,DBLP:conf/nips/ChenHYCSZ24} attempt to effectively optimize the target policies from multiple responses, while \citet{DBLP:journals/corr/abs-2405-19107} proposes direct reward optimization on single-trajectory data.
Other works focus on preference corruption~\citep{wu2025towards,DBLP:conf/icml/ChowdhuryKN24} or the preference distribution shift problem~\citep{DBLP:journals/corr/abs-2502-01930}.
From the perspective of modeling mechanism, recent techniques like token-level optimization~\citep{DBLP:conf/icml/ZengLMYZW24,DBLP:journals/corr/abs-2411-19943}, reference-free methods~\citep{DBLP:conf/emnlp/HongLT24,DBLP:conf/icml/XuSCTSDM024,DBLP:conf/nips/0001X024}, self-play optimization~\citep{DBLP:journals/corr/abs-2404-03715,DBLP:conf/icml/SwamyDK0A24,wu2025selfplay} exhibit practical potential.
It should be noted, however, that these methods generally underperform in reward modeling itself compared to explicit optimization results~\citep{DBLP:journals/corr/abs-2403-13787}.


\subsubsection{Reward Granularity Level} \label{Reward_Granularity}
In this subsection, we classify reward mechanisms based on their granularity when functioning as verifiers to solve problems with ground truth. Specifically, an Outcome-level Reward Model (ORM) predicts the probability that a completion results in a correct answer, while a Process-level Reward Model (PRM) assigns a score to each step in the reasoning process.

\paragraph{Outcome Level Reward.}
For tasks that require more complex reasoning, an ORM can be employed~\citep{starling2023,DBLP:journals/corr/abs-2410-18451,DBLP:conf/nips/YangDLZZ24}. Typically, the training data for an ORM is constructed differently from standard preference tuning~\citep{19ff28b9-64f9-3656-ba40-08326a05748e}. Specifically, each solution $s$ is paired with a problem statement or prompt $p$. The inductive bias appliedin this setup assumes that one completion represents a solution based on whether it is correct or not for the given problem.  The ORM $(P \times S \rightarrow \mathbb{R})$ is usually trained using a corss-entropy loss~\citep{li-etal-2023-making, cobbe2021training} 
\begin{equation}
\mathcal{L}_{ORM} = -( \hat{y_s} \log{y_s} + (1 - \hat{y_s}) \log{(1-y_s)})
\end{equation}

\paragraph{Process Level Reward.}

Despite their proficiency in multi-step reasoning tasks, outcome-supervised methods are still prone to hallucinations, such as reaching the correct answer through an incorrect reasoning path~\citep{DBLP:conf/iclr/CreswellSH23}. This indicates the necessity of incorporating process supervision to address these limitations. Additionally, the PRM $(P \times S \rightarrow \mathbb{R^{+}} )$ can be trained using the standard classification loss function below, where $y_i$ is the prediction score of the PRM and $\hat{y_i}$ represents the correctness label, and $N$ is the total number of reasoning step for $s$.
\begin{equation}
\mathcal{L}_{PRM} = - \sum_{i=1}^{N} \hat{y_{s_i}} \log{y_{s_i}} + (1 - \hat{y_{s_i}}) \log{(1-y_{s_i})} 
\end{equation}


Different from the heuristic method~\citep{li-etal-2023-making} that leverage semantically relevance for stepwise verification,
\citet{DBLP:journals/corr/abs-2211-14275} introduces the PRM which are trained on human annotated stepwise labels. The PRM evaluates each reasoning step individually and can reduce tracking error and avoid tampering incentives~\citep{DBLP:conf/ijcai/EverittKOL17}.
Moreover, \citet{DBLP:conf/iclr/LightmanKBEBLLS24} constructs a large-scale stepwise human feedback dataset PRM800K and train a PRM to predict step correctness in the form of tokens.

To further reduce the cost of human annotation,
\citet{wang2024math} and \citet{DBLP:conf/emnlp/WangLWLH0S24} 
obtain process-supervised signals based on the Monte Carlo (MC) method. For each step and prefix, the frequency of reaching correct answer within sampled completions can be used to estimate step quality, i.e., Q-value function~\citep{DBLP:journals/corr/MnihKSGAWR13}.
Expanding on them, \citet{DBLP:journals/corr/abs-2406-06592} employs an adaption of Monte Carlo Tree Search (MCTS) to construct state-action trees for collecting the PRM training data.
In addition, \citet{DBLP:conf/emnlp/KhalifaLLL023} proposes a stepwise discriminator through contrastive learning, where the preference pairs are obtained by aligning LLM-generated incorrect solutions with the reference solution.

Another series of works argue process reward should measure progress and advantages. \citet{DBLP:journals/corr/abs-2410-11287} frames the PRM as a ranking problem to capture inter-dependencies among reasoning steps, which means larger Q-value is more likely to reach correct answer, and a significant gap will exist between correct steps and the first incorrect step. \citet{DBLP:conf/nips/LuD0CDFG24} and \citet{DBLP:journals/corr/abs-2410-08146} introduce advantages as PRMs to measure the confidence and likelihood change of generating a correct response. Furthermore, \cite{DBLP:journals/corr/abs-2412-01981} and \citet{cui2025process} obtains implicit PRMs from trained ORMs through reward parameterization, which can be used to estimate advantages by calculating  token-level rewards.

\paragraph{Pros and Cons of different types of RMs.} 

\begin{table*}[t] 
    \small
    \caption{Comparison of advantages and disadvantages of the different granularity of RMs}
    \renewcommand{\arraystretch}{1.5}
        \begin{tabular}{lll}
        \toprule[1pt]
        \makebox[0.12\linewidth][l]{\textbf{Granularity}} & \makebox[0.33\linewidth][l]{\textbf{Advantages}} & \makebox[0.33\linewidth][l]{\textbf{Disadvantages}} \\ \hline
        \textbf{Outcome}        & Potential in flexible tasks & Leading to false positives solutions. \\ 
                                & Ease of implementation & Sparse reward \\ 
                                \hline
        \textbf{Process}        & Potential in reasoning tasks. & High cost for gathering training data. \\ 
                                & Dense reward & Value estimation yields inferior performance. \\
                                & Controllable & Hard to define process reward. \\
                                &  & Scalability and generalization problems \\
        \bottomrule[1pt]
        \end{tabular}
    
    \label{table:granularity}
\end{table*}

Currently, ORM tends to be better than PRM in the tasks with flexible processes due to its ease of implementation and generalizability, but it may lead to false positives solutions~\citep{DBLP:conf/iclr/CreswellSH23} in the reasoning tasks. 
PRM has demonstrated its potential in reasoning tasks~\citep{DBLP:journals/corr/abs-2310-10080,DBLP:journals/corr/abs-2406-06592}, but there are several considerations that require attention.
Manual annotation is expensive and not scalable~\citep{song2025prmbench}, while automated annotation may not produce satisfactory results. 
\citet{prmlessons} finds that MC estimation hinder the capability of PRMs to identify incorrect steps compared to judge LLMs.
Besides, process rewards are difficult to define~\citep{cui2025process}, determining the correctness of intermediate steps and the progress of solving problems is challenging.
Moreover, it is often suffers from reward hacking~\citep{wang2025hierarchicalmultisteprewardmodels}, while retraining the RM introduces additional complexity and resource requirements.
Finally, although PRM excels at reranking top-N responses or assisting in guided search~\citep{snell2024scaling}, its computational overhead in large-scale reinforcement learning tasks outweighs its benefits in practical experiments~\citep{deepseekai2025deepseekr1incentivizingreasoningcapability}.
An overview of the opinions is in Table~\ref{table:granularity}.

\subsection{Usage}
\label{2_3_usage}

In the context of LLMs, RMs serve as critical components that help guide model behavior toward desired outcomes. By defining a structured, quantifiable signal that measures how well a generated response aligns with specific goals or user preferences, RMs enable the tuning and optimization of LLM outputs. This RM utility manifests across multiple stages of the LLM life cycle, including data selection, policy training, and the inference stage. In this subsection, we investigate RM utility from these three perspectives in detail. 

\paragraph{Data Selection}
Some studies utilize RMs to select data for the fine-tuning of LLMs. \citet{DBLP:journals/tmlr/Dong0GZCPDZS023} proposes an SFT-like iterative training method, where an RM is utilized to rank the quality of LLM-generated responses. Data with the highest reward can be used to finetune the LLM. \citet{DBLP:conf/nips/YuanYTWHH23} further introduces ranking loss to align the LLM-generated score with the RM-generated score. \citet{DBLP:journals/corr/abs-2308-08998} leverages an RM-filtered dataset to fine-tune LLM towards an offline RL objective. \citet{DBLP:conf/nips/PangYHCSW24} evaluates answers and rationale for correctness by RMs, thereby selecting peference pairs to optimize LLMs via DPO~\citep{rafailov2023direct} objective.

\paragraph{Policy Training.} RMs provide feedback signals that reinforce or penalize certain behaviors~\citep{ouyang2022training}, ultimately shaping the model’s decision-making policies. To mitigate the issue of low robustness, which arises primarily because the RM often struggles with out-of-distribution generalization~\citep{DBLP:journals/corr/abs-2311-14743} and mismatched human judgment, several strategies have been investigated. These include length-controlled reward setting~\citep{chen2024odin,DBLP:conf/cncl/ZhouWHXZZ24,DBLP:conf/acl/ParkREF24}, causal reward modeling~\citep{DBLP:journals/corr/abs-2501-09620,liu2025rrm}, Bayesian method~\citep{yang2024bayesian,DBLP:journals/corr/abs-2402-09764,yan2024rewardrobustrlhfllms}, and ensemble~\citep{wu2024fine,ramé2024warmbenefitsweightaveraged,zhang2024improvingreinforcementlearninghuman}. 

\paragraph{Inference.} RMs can be used to rank multiple outputs to deliver responses that best align with application-specific criteria. As discussed in \S\ref{Reward_Granularity}, RMs can be classified as ORM and PRM. PRMs are often used at the inference stage to evaluate the progress and improve reasoning ability~\citep{DBLP:journals/corr/abs-2410-08146}. Some RM-guided tree search frameworks~\citep{DBLP:journals/corr/abs-2310-10080,jiang2024technical,DBLP:conf/emnlp/He0ZT024,DBLP:conf/nips/ZhangZHYD024} which have been shown to be able to greatly enhance the reasoning abilities of LLMs. In addition, RMs can also be used to evaluate intermediate decoding steps and dynamically decide whether to invoke a more powerful target model to balance resource utilization and performance~\citep{liao2025reward}.

\section{Applications}
\label{4_application}

RMs have found extensive applications across multiple domains. Here, we briefly summarize some key areas where RMs are currently utilized. 

\paragraph{Dialogue.}
RMs help mitigate harmful responses by refining them based on ethical guidelines and user intent~\citep{bai2022training,glaese2022improving,DBLP:journals/corr/abs-2212-08073,DBLP:conf/iclr/DaiPSJXL0024,DBLP:journals/corr/abs-2412-16339}. Meanwhile, some works focus on the professionalism~\citep{zhang2023huatuogpt,yang2024zhongjing} in dialogue, requiring agents to accurately and clearly express complex knowledge.
Other works attempt to improve the overall dialogue impression~\citep{DBLP:journals/corr/abs-2408-02976,DBLP:journals/corr/abs-2501-12698}, including empathy, enthusiasm, humanlikeness, and so on.

\paragraph{Reasoning.}
In mathematical reasoning~\citep{DBLP:journals/corr/abs-2110-14168,DBLP:journals/corr/abs-2211-14275}, RMs, especially PRM, can provide guidance to LLMs to improve logical consistency by balancing the exploration of various solutions with minimizing errors~\citep{DBLP:journals/corr/abs-2308-09583,DBLP:journals/corr/abs-2406-06592,wang2024math,DBLP:conf/iclr/LightmanKBEBLLS24,shao2024deepseekmath,zhang2025lessons,zhu2025retrievalaugmentedprocessrewardmodel}. Additionally, RMs have also shown promise in code generation~\citep{DBLP:journals/corr/abs-2412-20367} by integrating API calls, improving learning efficiency, and optimizing performance~\citep{DBLP:journals/corr/abs-2310-10080,DBLP:journals/corr/abs-2406-20060,DBLP:journals/corr/abs-2410-17621,nichols2024performance,DBLP:journals/corr/abs-2409-06957,mcaleese2024llm}.

\paragraph{Retrieve \& Recommendation.}
RMs can be employed to help align the retrieve process with the preferences of strong LLMs
~\citep{xiong2024search}, which include assessing relevance~\citep{DBLP:conf/emnlp/0002DW23, kim2025syntriever}, adaptive retrieval~\citep{guan2025deepragthinkingretrievalstep}, and improving the quality of intermediate queries~\citep{xiong2025raggymoptimizingreasoningsearch}. As for recommendation systems, RMs can be used to capture nuanced user preferences~\citep{DBLP:conf/sigir/WangKAJ24}, evaluate LLM-generated user preferences~\citep{DBLP:journals/corr/abs-2410-05939}, and lead to
high-quality explanations~\citep{DBLP:conf/aaai/0002ZWCZ0HZY24}.

\paragraph{Other Applications.}
Apart from aforementioned applications in the text domain, RMs have demonstrated potential in other modalities, such as text to audio~\citep{DBLP:conf/icml/CideronGVVKBMUB24, DBLP:conf/ijcai/LiaoHYD0XXLL024,DBLP:journals/corr/abs-2405-14632}, text to image~\citep{DBLP:journals/corr/abs-2302-12192,DBLP:conf/nips/XuLWTLDTD23,DBLP:journals/corr/abs-2305-16381}, text to video~\citep{DBLP:conf/nips/WuHWXW24a,DBLP:conf/cvpr/YuanZW0FPZ0A024,wang2025harnesslocalrewardsglobal}. Moreover, RMs have been explored in some interactive tasks including robotic manipulation~\citep{DBLP:journals/corr/abs-2311-02379,DBLP:conf/iclr/RocamondeMNPL24} and games~\citep{DBLP:journals/corr/abs-2307-12158,choudhury2025processrewardmodelsllm}, which become the foundation of artificial general intelligence.

\section{Benchmarks}
\label{5_evaluation}

RM evaluation is crucial because errors in RM can negatively affect the performance of the final policy~\citep{DBLP:journals/corr/abs-2410-14872, DBLP:journals/corr/abs-2410-05584, DBLP:journals/corr/abs-2407-18369}. 
However, the development of general and standardized benchmarks for RM evaluation remains nascent, making it hard to compare and improve RMs. 
This is due to several challenges: (1) The most direct way to evaluate an RM is to train a full RL policy and observe its performance, which is very costly~\citep{DBLP:journals/corr/abs-2410-14872}. (2) RM evaluation is often tied to the performance of the policy trained with it, making it difficult to assess the RM independently~\citep{DBLP:journals/corr/abs-2403-13787}. (3) While creating a dataset for evaluation (e.g., annotating a simple pairwise comparison dataset) is relatively easy, RMs are sensitive to changes in input style, domain, or format~\citep{DBLP:journals/corr/abs-2410-16184}.
This means RM evaluation requires a more comprehensive approach, considering constructing more dynamic, multi-faceted testing, which further compounds the difficulty. Recently, researchers have tried to construct high-quality benchmarks to explore optimizing RMs within different RL policies, LM architectures, training budgets, etc.


\paragraph{ORM Benchmarks.} \citet{DBLP:journals/corr/abs-2403-13787} constructs a comprehensive benchmark \texttt{RewardBench}, which contains human-verified prompt-chosen-rejected trios spanning chat, reasoning, safety, and prior test sets, meanwhile providing a toolkit to audit RM behavior. \citet{DBLP:journals/corr/abs-2410-16184} proposes \texttt{RM-Bench}, which includes chat, code, math, and safety annotated data, and conducts large-scale evaluation on publicly accessible RMs. \citet{DBLP:journals/corr/abs-2410-09893} introduces \texttt{RMB} that involves over 49 real-world scenarios, and discusses the generalization defects in previous benchmarks. 
Specifically, \citet{DBLP:journals/corr/abs-2410-14872} proposes \texttt{PPE} that evaluate RMs on proxy tasks (related to downstream RLHF outcomes) by launching an end-to-end RLHF experiment.

\paragraph{PRM Benchmarks.} With the emergence of reasoning research, LMs are adapted to more complex scenarios like math and multi-hop decision-making tasks, therefore PRMs have appeared and been applied. For evaluating PRMs, \citet{DBLP:journals/corr/abs-2412-06559} propose \texttt{ProcessBench}, which consists of a huge number of cases with annotated step-by-step solutions on competition math problems.
\citet{song2025prmbench} introduce \texttt{PRMBench}, comprises thousands of designed problems with stepwise labels, evaluating RMs across multiple dimensions.

In addition to aforementioned studies, some recent works evaluate RMs for specific domains or applications, e.g., Vision-Language~\citep{DBLP:journals/corr/abs-2411-17451,DBLP:journals/corr/abs-2407-04842,yasunaga2025multimodalrewardbenchholisticevaluation}, Multilingual Settings~\citep{DBLP:journals/corr/abs-2410-15522}, and Retrieve-Augmented Generation~\citep{Jin2024RAGRewardBenchBR}.
These benchmarks collectively mitigate the need for a more comprehensive and fine-grained evaluation of RMs, paving the way for more reliable and robust RMs for training stronger LMs.

\section{Challenges}
\label{6_challenges}

\subsection{Data}
High-quality data collection to reflect human preference is the cornerstone of RM applications, but there still remains some challenges associated with its process.
During the collection, potential biases may exist between the preferences of researchers and annotators~\citep{DBLP:journals/corr/abs-1909-08593}. 
Variations in expertise among annotators can introduce noisy data~\citep{DBLP:journals/corr/abs-2211-06519,DBLP:conf/aaai/BarnettFS023}, which may be significant in some knowledge-intensive tasks. 
The issue of assessment quality can result in the inconsistencies~\citep{DBLP:conf/iclr/BansalDG24} between sparse feedback protocols (i.e., ratings and rankings), when dense feedback is expensive to collect.
To tackle above challenges, data filtering~\citep{DBLP:journals/corr/abs-2406-16486}, selection~\citep{DBLP:journals/corr/abs-2502-14560}, and high-fidelity scalable synthetic data~\citep{DBLP:journals/corr/abs-2404-07503} become promising solutions.

\subsection{Training}
A critical challenge in RM training is overoptimization which also called reward hacking~\citep{DBLP:journals/corr/abs-2209-13085,gao2023scaling,weng2024rewardhack,laidlawcorrelated}, where RMs could be excessively optimized to narrow evaluation metric (like accuracy on a sole static benchmark)~\citep{DBLP:journals/corr/abs-2410-05584}.
An RL policy trained against such RMs may ``hack'' the reward signal, leading to performance degradation~\citep{DBLP:journals/corr/abs-2009-01325}. 
Some causes of overoptimization include reward tampering~\citep{mckee2024honesty,DBLP:journals/corr/abs-2406-10162}, mislead~\citep{DBLP:journals/corr/abs-2409-12822}, and sycophancy~\citep{DBLP:conf/iclr/SharmaTKDABDHJK24}.
As mentioned in \S\ref{2_3_usage}, several research directions such as RM ensemble~\citep{coste2024rewardmodelensembleshelp}, data augmentation~\citep{liu2025rrm}, and robust training~\citep{laidlawcorrelated,zhang2024overcomingrewardoveroptimizationadversarial,miao2025energylossphenomenonrlhf} have demonstrated potential in mitigating overoptimization, paving the way for more robust RMs. 

\subsection{Bias in Evaluation}

Using RMs (judge model) for evaluation also introduces intrinsic biases toward superficial quality of text~\citep{DBLP:journals/corr/abs-2403-02839}.
\citet{DBLP:journals/corr/abs-2409-11704} observe that top-ranking RMs and some popular benchmarks exhibit biases toward the specific format patterns
\citet{park2024offsetbias} discuss the biases derived from evaluators, including length, concreteness, empty reference, and so on.
\citet{DBLP:journals/corr/abs-2502-01534} study the preference leakage problem elicited by the relevance between synthetic data generators and RMs.
The aforementioned studies highlight the need to construct robust evaluation benchmarks to detect and mitigate biases.

\section{Future Directions}
\label{7_future_directions}

\paragraph{The combination of scalar rewards with rule-based rewards is becoming a growing trend.}
In advanced industrial LLMs~\citep{deepseekai2025deepseekr1incentivizingreasoningcapability, liu2024deepseek}, a robust model can benefit from integrating rule-based and model-based rewards. Rule-based rewards provide clear guidelines, while model-based rewards enable learning from predictions. 
Specifically, rule-based rewards are applied to tasks with clear ground truths (e.g., mathematics, coding), while reward models are used for tasks without clear ground truths (e.g., creative tasks), enhancing LLMs' real-world applicability. Incorporating rule-based rewards has become a standard practice in the reinforcement fine-tuning of o1-like~\citep{jaech2024openai} longCoT models, and a few works~\citep{tinyzerO, OpenR1, OpenThoughts} in the academic community which only utilize rule-based reward have emerged, also achieving strong reasoning capabilities.

\paragraph{Reward Design in LLM Long-horizontal Agent Task.}
Recent advances in reasoning ability have enabled sophisticated LLMs to tackle complex expert-level tasks~\citep{phan2025humanitysexam}, with planning playing a key role. OpenAI and Anthropic are exploring tool use, such as search engines~\citep{deepresearch2025}, code interpreters~\citep{cursor2025}, and web browsers~\citep{operator2025} to complete complex GUI tasks~\citep{zhang2024large}. However, ensuring good agent performance is challenging, especially when designing feedback mechanisms for large systems. Creating rules is experimental, and developing an end-to-end reinforcement learning framework for long-horizontal tasks is essential. The key challenge remains ensuring the agent consistently receives rewards and improves monotonically.

\paragraph{Empowering the multi-modal domain.}
RMs are rapidly evolving in the multi-modal domain, which includes the integration of modalities such as image, audio, and video. Compared to single-modality, the collection of multi-modal preference data is more costly. Some techniques such as few-shot learning~\citep{DBLP:conf/corl/HejnaS22}, data synthesis~\citep{DBLP:journals/corr/abs-2412-17417} remain to be explored, thereby reducing the reliance on human annotators. Meanwhile, designing a high-quality reward signal~\citep{Narin2024EvolutionaryRD} is crucial, which involves alignment across different modalities. Finally, exploring methods to enhance the cross-domain generalization of RMs, and bridging the gap between simulated and real-world scenarios, will contribute to the realization of embodied intelligence.

\section{Conclusion and Discussion}

In this paper, we present the first comprehensive survey specifically focused on Reward Models in the LLM era. We systematically review related studies of RMs, introduce an elaborate taxonomy, discuss the practical applications, highlight the challenges, and explore potential research directions. 
Besides, we discuss some open questions about RMs. (1) Is Rule-based reward enough for RL? (2) Is Mixture-of-Experts better than BT Model? (3) How to overcome the reward hacking of RM as LLMs surpass the level of the best expert level? See Appendix \ref{open_questions} for more details.
We hope that this survey will be helpful to researchers and facilitate further research.

\bibliography{colm2025_conference}
\bibliographystyle{colm2025_conference}

\appendix
\section{Appendix}
\label{sec:appendix}

\subsection{Relevant Survey}
\label{related_surey}

Some previous surveys focus on human-involved RL~\citep{DBLP:conf/ACMdis/CruzI20,DBLP:journals/firai/NajarC21,DBLP:journals/jair/RetzlaffDWMAYSATH24}, while~\citet{10766898} discusses LLM-enhanced RL.
\citet{DBLP:journals/corr/abs-2310-19852} and~\citet{DBLP:journals/corr/abs-2407-16216} conducts a comprehensive investigation on LLM alignment.
\citet{DBLP:journals/corr/abs-2312-14925} and~\citet{DBLP:journals/tmlr/CasperDSGSRFKLF23} both focus on RLHF, while~\citet{DBLP:journals/corr/abs-2312-14925} discusses the researches in which RM is the sore source of information for the objective.~\citet{DBLP:journals/tmlr/CasperDSGSRFKLF23} overviews the open problems and limitations of RLHF.

Compared with the aforementioned survey, our work primarily focuses on RMs in LLM era. We systematically introduce RMs based on their life-cycles, and explain the popular usages and evaluation perspectives. In addition, we discuss the challenges and potential research directions of RMs in detail. We sincerely hope that this paper can deepen researchers' understanding of the field and facilitate future works.


\subsection{Reward Modeling}

The Bradley-Terry Model~\citep{19ff28b9-64f9-3656-ba40-08326a05748e} can be used for modeling pairwise preference, which is the most commonly reward model assumption. For a prompt $x$, reward model $r$, response pair $y_w,y_l$. It estimates the probability of prefer to $y_w$ rather than $y_l$:

$$P(y_w \succ y_l|x) = \frac{1}{1+exp(r(x,y_w)-r(x,y_l))}.$$

An RM $\widehat{r}$ can be derived by optimizing the following maximum likelihood objectives, where $\mathcal{D}$ and $\sigma$ represent the preference dataset and sigmoid function respectively.

$$
\widehat{r}\leftarrow\mathop{\arg\max}_{r\in\mathcal{R}}\mathbb{E}_{(x,y_w,y_l)\sim\mathcal{D}}\left[\log\sigma(r(x,y_w)-r(x,y_l))\right].
$$

Under RLHF setting~\citep{ouyang2022training}, the target policy model is optimized by using the learned RM $\widehat{r}(x,y)$. $\pi_{\mathsf{ref}}(x,y)$ represent the reference model before update, and the resulting Kullback-Leibler (KL) penalty term is utilized to constrain the size of the policy update~\citep{DBLP:journals/corr/SchulmanWDRK17}:

$$
\widehat{\pi}\leftarrow\mathop{\arg\max}_{\pi\in\Pi}\mathbb{E}_{x\sim\mathcal{D},y\sim\pi(\cdot|x)}\left[\widehat{r}(x,y)-\beta\log\frac{\pi(x,y)}{\pi_{\mathsf{ref}}(x,y)}\right]
$$

DPO~\citep{rafailov2023direct} is an alternative alignment approach which can optimize the policy without explicit reward modeling:
$$
\widehat{\pi}\leftarrow
\mathop{\arg\max}_{\pi\in\Pi}\mathbb{E}_{(x,y_w,y_l)\sim\mathcal{D}}\left[\log\sigma\left(\beta\log\frac{\pi(y_w\mid x)}{\pi_{\mathrm{ref}}(y_w\mid x)}-\beta\log\frac{\pi(y_l\mid x)}{\pi_{\mathrm{ref}}(y_l\mid x)}\right)\right],
$$

where $\beta$ is a scalable parameter.


\subsection{Reward Shaping \& Ensemble} 
\label{3_ensemble}

A major challenge in real-world scenarios is the sparsity and delay of rewards, which can hinder learning. This section focuses on engineering the reward model~\citep{kwon2023rewarddesignlanguagemodels} during reinforcement learning .

\paragraph{Reward on Point-wise Feedback}
Pointwise feedback assigns numerical values to actions or outcomes, enabling precise adjustments to the agent's policy. It is effective for tasks where each action’s quality can be independently assessed. For example, \citet{pace2024west} and \citet{jinnai2024regularized} propose a self-training strategy to select the best and worst reward samples. \citet{wang2024secretsrlhflargelanguage} addresses ambiguous preference pairs by incorporating a margin in the reward, improving model generalization. \citet{liu2024skyworkrewardbagtricksreward} employs a data-centric approach to enhance feedback quality and make reward models more effective. 

\paragraph{Reward on Binary Feedback}
Binary feedback simplifies evaluation by categorizing outcomes as positive or negative, eliminating the need for a ground truth. This makes implementation and interpretation easier. For instance, Nash learning~\citep{munos2024nashlearninghumanfeedback} models pairwise preferences by binary feedback but struggles with inconsistent human labeling. Approaches like KTO~\citep{ethayarajh2024ktomodelalignmentprospect} use the Kahneman-Tversky model~\citep{levy1992introduction} to maximize utility, and DRO~\citep{richemond2024offlineregularisedreinforcementlearning} combines offline reinforcement learning with regularization in binary feedback. Binary feedback also guides agent learning by signaling desirable actions, as explored in \citet{wachi2024longtermsafereinforcementlearning}. However, it may not capture the full complexity of human preferences. 


\paragraph{Reward on Ensemble Feedback}
Model ensemble~\citep{Ganaie_2022} is a classic machine learning method for mitigating reward overoptimization and improving policy optimization. Typically, ensemble feedback~\citep{ramé2024warmbenefitsweightaveraged, coste2024rewardmodelensembleshelp, pace2024west, wu2024fine} aims to combine reward signal to further reduce reward hacking during reinforcement fine-tuning. For computational efficiency, \citet{zhang2024improvingreinforcementlearninghuman} propose a LoRA-based ensemble method that reduces the computational cost associated with reward ensembles. Additionally, reward ensemble techniques, such as the Bayesian ensemble method~\citep{yan2024rewardrobustrlhfllms}, can be used to approximate uncertainty in the feedback.


\subsection{Open Questions}
\label{open_questions}

\paragraph{Is Rule-based reward enough for RL?}

Rule-based rewards are a good way to mitigate reward hacking, but it’s hard to say whether they are enough on their own. Without sufficient supervision, large language models (LLMs) may encounter very sparse rewards, leading to optimization divergence. Additionally, for tasks that don’t have a clear ground truth, designing an effective rule-based reward can be challenging. In such cases, preference learning can be a better option, as it allows us to derive reward signals from comparative feedback rather than relying solely on predefined rules. Thus, while rule-based rewards can be helpful, they may not always provide the necessary robustness for complex tasks.


\paragraph{Is Mixture-of-Experts better than BT Model?} 
There are several works related to Mixture-of-Experts (MoE) models, such as the DMoERM model \cite{quan2024dmoermrecipesmixtureofexpertseffective} and LoRA-ensemble \cite{halbheer2024loraensembleefficientuncertaintymodelling, dou2024loramoealleviateworldknowledge}. MoE models have shown great potential in creating Pareto-optimal \cite{lee2024parrotparetooptimalmultirewardreinforcement, rame2024rewarded} reward models, where they can combine multiple expert models to focus on different aspects of the problem, offering a more versatile and efficient approach. While the BT model has its strengths, MoE models have the advantage of scalability and the ability to improve performance by selecting the most relevant expert for each situation. This flexibility often leads to better generalization and optimization, especially in complex tasks.

\paragraph{How to overcome the reward hacking of RM as LLMs surpass the level of best expert level?}
As LLMs surpass the performance of the best expert models, overcoming reward hacking becomes more challenging. One approach is to shift from weak-to-strong generalization \cite{burns2023weak}. This involves designing reward models that encourage more robust, flexible learning that accounts for a wider variety of potential behaviors and outcomes. Instead of relying solely on expert-level feedback, incorporating broader, more generalized reward signals helps ensure that the system doesn't exploit narrow solutions or hacks. This strategy promotes more meaningful generalization and prevents the model from exploiting loopholes in the reward structure.


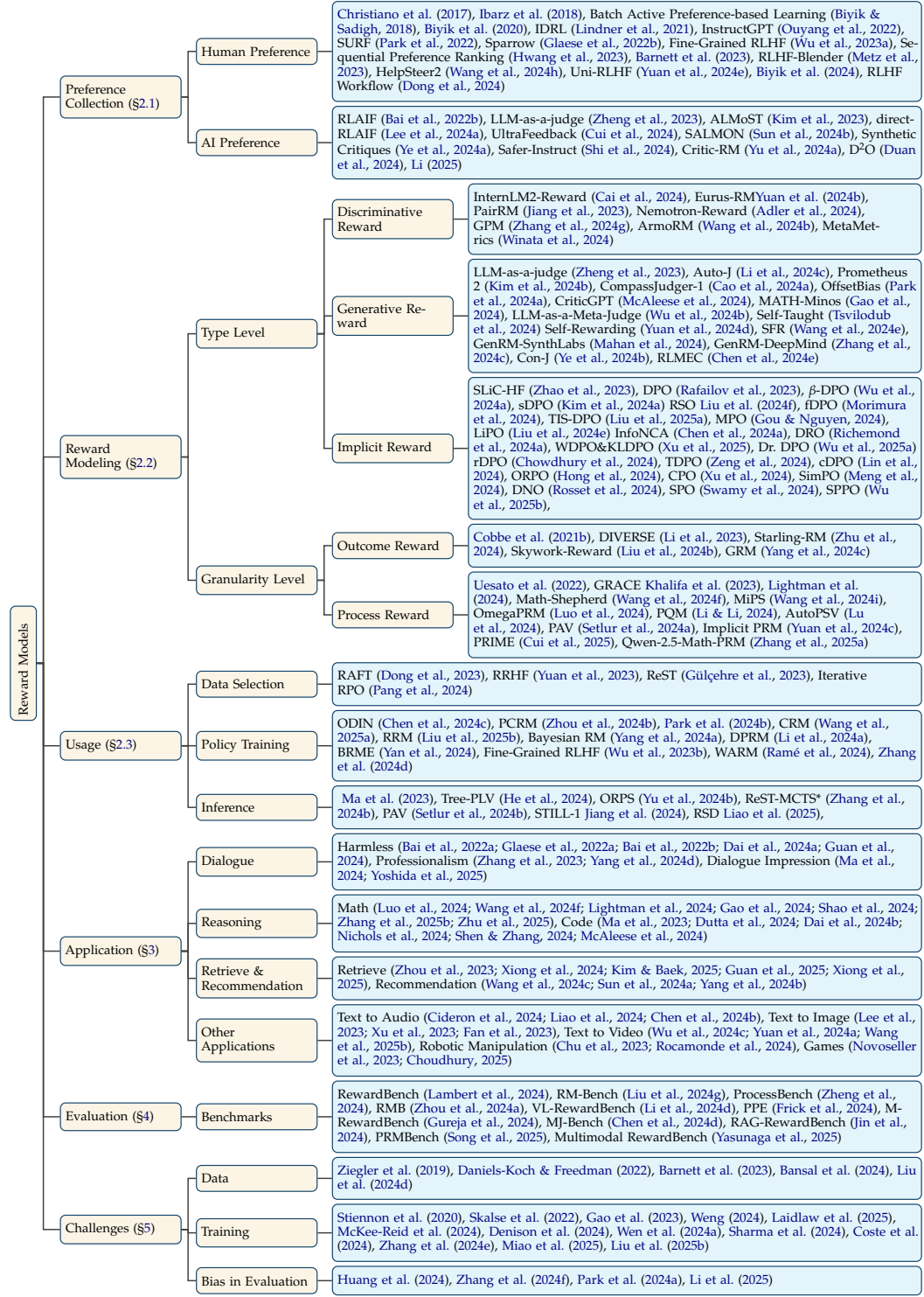
\begin{figure*}[t]
    \vspace{-2mm}
    \centering
    \resizebox{\textwidth}{!}{
        \begin{forest}
            forked edges,
            for tree={
                child anchor=west,
                parent anchor=east,
                grow'=east,
                anchor=west,
                base=left,
                font=\normalsize,
                rectangle,
                draw=hidden-black,
                rounded corners,
                minimum height=2em,
                minimum width=4em,
                edge+={darkgray, line width=1pt},
                s sep=3pt,
                inner xsep=0.4em,
                inner ysep=0.6em,
                line width=0.8pt,
                text width=8.5em,
                ver/.style={
                    rotate=90,
                    child anchor=north,
                    parent anchor=south,
                    anchor=center,
                    text width=8em
                },
                node/.style={
                    fill opacity=.2,
                    fill=hidden-orange!90
                },
                leaf/.style={
                    text opacity=1,
                    inner sep=2pt,
                    fill opacity=.5,
                    fill=hidden-blue!90, 
                    text=black,
                    text width=44.5em
                    font=\normalsize,
                    inner xsep=0.4em,
                    inner ysep=0.6em,
                    draw,
                }, 
            },
            [
                Reward Models, ver, node
                [
                    Preference \\Collection~(\S\ref{2_1_preference_collection}), node
                    [
                        Human Preference, node
                        [
                            \citet{DBLP:conf/nips/ChristianoLBMLA17}{,}
                            \citet{DBLP:conf/nips/IbarzLPILA18}{,}
                            Batch Active Preference-based Learning~\citep{DBLP:conf/corl/BiyikS18}{,}
                            \citet{DBLP:conf/rss/BiyikHKS20}{,}
                            IDRL~\citep{DBLP:conf/nips/LindnerTTCK21}{,}
                            InstructGPT~\citep{ouyang2022training}{,}
                            SURF~\citep{DBLP:conf/iclr/ParkSSLAL22}{,}
                            Sparrow~\citep{DBLP:journals/corr/abs-2209-14375}{,}
                            Fine-Grained RLHF~\citep{DBLP:conf/nips/WuHSDSASOH23}{,}
                            Sequential Preference Ranking~\citep{DBLP:conf/nips/HwangLKK0O23}{,}
                            \citet{DBLP:conf/aaai/BarnettFS023}{,}
                            RLHF-Blender~\citep{DBLP:journals/corr/abs-2308-04332}{,}
                            HelpSteer2~\citep{DBLP:journals/corr/abs-2406-08673}{,}
                            Uni-RLHF~\citep{DBLP:conf/iclr/YuanHMDL0FZ024}{,}
                            \citet{DBLP:journals/thri/BiyikAS24}{,}
                            RLHF Workflow~\citep{DBLP:journals/corr/abs-2405-07863}
                            , leaf, text width=44.6em
                        ]
                    ]
                    [
                        AI Preference, node
                        [
                            RLAIF~\citep{DBLP:journals/corr/abs-2212-08073}{,}
                            LLM-as-a-judge~\citep{zheng2023judging}{,}
                            ALMoST~\citep{DBLP:conf/emnlp/KimBSKKYS23}{,}
                            direct-RLAIF~\citep{DBLP:conf/icml/0001PMMFLBHCRP24}{,}
                            UltraFeedback~\citep{DBLP:conf/icml/CuiY0YH0NXXL0024}{,}
                            SALMON~\citep{DBLP:conf/iclr/SunSZZCCYG24}{,}
                            Synthetic Critiques~\citep{DBLP:journals/corr/abs-2405-20850}{,}
                            Safer-Instruct~\citep{DBLP:conf/naacl/ShiCZ24}{,}
                            Critic-RM~\citep{DBLP:journals/corr/abs-2411-16646}{,}
                            D$^2$O~\citep{DBLP:conf/emnlp/DuanY0LLLXG24}{,}
                            \citet{decision-tree-perspective}
                            , leaf, text width=44.6em
                        ]
                    ]
                ]
                [
                    Reward \\Modeling~(\S\ref{2_2_reward_modeling}), node
                    [
                        Type Level, node
                        [
                            Discriminative \\Reward, node
                            [
                                InternLM2-Reward~\citep{DBLP:journals/corr/abs-2403-17297}{,}
                                Eurus-RM\cite{DBLP:journals/corr/abs-2404-02078}{,}
                                PairRM~\citep{DBLP:conf/acl/Jiang0L23}{,}
                                Nemotron-Reward~\citep{DBLP:journals/corr/abs-2406-11704}{,}
                                GPM~\citep{zhang2024general}{,}
                                ArmoRM~\citep{wang2024interpretablepreferencesmultiobjectivereward}{,}
                                MetaMetrics~\citep{DBLP:journals/corr/abs-2410-02381}
                                , leaf, text width=34.1em
                            ]
                        ]
                        [
                            Generative Reward, node
                            [
                                LLM-as-a-judge~\citep{zheng2023judging}{,}
                                Auto-J~\citep{li2023generative}{,}
                                Prometheus\ 2~\citep{kim2024prometheus}{,}
                                CompassJudger-1~\citep{DBLP:journals/corr/abs-2410-16256}{,}
                                OffsetBias~\citep{park2024offsetbias}{,}
                                CriticGPT~\citep{mcaleese2024llm}{,}
                                MATH-Minos~\citep{DBLP:journals/corr/abs-2406-14024}{,}
                                LLM-as-a-Meta-Judge~\citep{wu2024metarewardinglanguagemodelsselfimproving}{,}
                                Self-Taught~\citep{tsvilodub2024towards}
                                Self-Rewarding~\citep{yuan2024selfrewardinglanguagemodels}{,}
                                SFR~\citep{wang2024direct}{,}
                                GenRM-SynthLabs~\citep{mahan2024generativerewardmodels}{,}
                                GenRM-DeepMind~\citep{zhang2024generativeverifiersrewardmodeling}{,}
                                Con-J~\citep{ye2024scalarrewardmodellearning}{,}
                                RLMEC~\citep{DBLP:conf/acl/ChenZZWZZW24}
                                , leaf, text width=34.1em
                            ]
                        ]
                        [
                            Implicit Reward, node
                            [
                                SLiC-HF~\citep{DBLP:journals/corr/abs-2305-10425}{,}
                                DPO~\citep{rafailov2023direct}{,}
                                $\beta$-DPO~\citep{DBLP:conf/nips/WuX0WGDW024}{,}
                                sDPO~\citep{DBLP:journals/corr/abs-2403-19270}
                                RSO~\citet{DBLP:conf/iclr/0002ZJKSLL24}{,}
                                fDPO~\citep{DBLP:conf/emnlp/MorimuraSJAA24}{,}
                                TIS-DPO~\citep{liu2025tisdpo}{,}
                                MPO~\citep{DBLP:journals/corr/abs-2403-19443}{,}
                                LiPO~\citep{DBLP:journals/corr/abs-2402-01878}
                                InfoNCA~\citep{DBLP:conf/nips/ChenHYCSZ24}{,}
                                DRO~\citep{DBLP:journals/corr/abs-2405-19107}{,}
                                WDPO\&KLDPO~\citep{DBLP:journals/corr/abs-2502-01930}{,}
                                Dr.\ DPO~\citep{wu2025towards}
                                rDPO~\citep{DBLP:conf/icml/ChowdhuryKN24}{,}
                                TDPO~\citep{DBLP:conf/icml/ZengLMYZW24}{,}
                                cDPO~\citep{DBLP:journals/corr/abs-2411-19943}{,}
                                ORPO~\citep{DBLP:conf/emnlp/HongLT24}{,}
                                CPO~\citep{DBLP:conf/icml/XuSCTSDM024}{,}
                                SimPO~\citep{DBLP:conf/nips/0001X024}{,}
                                DNO~\citep{DBLP:journals/corr/abs-2404-03715}{,}
                                SPO~\citep{DBLP:conf/icml/SwamyDK0A24}{,}
                                SPPO~\citep{wu2025selfplay}{,}
                                , leaf, text width=34.1em
                            ]
                        ]
                    ]
                    [
                        Granularity Level, node
                        [
                            Outcome Reward, node
                            [
                                \citet{cobbe2021training}{,}
                                DIVERSE~\citep{li-etal-2023-making}{,}
                                Starling-RM~\citep{starling2023}{,}
                                Skywork-Reward~\citep{DBLP:journals/corr/abs-2410-18451}{,}
                                GRM~\citep{DBLP:conf/nips/YangDLZZ24}
                                , leaf, text width=34.1em
                            ]
                        ]
                        [
                            Process Reward, node
                            [
                                \citet{DBLP:journals/corr/abs-2211-14275}{,}
                                GRACE~\citet{DBLP:conf/emnlp/KhalifaLLL023}{,}
                                \citet{DBLP:conf/iclr/LightmanKBEBLLS24}{,}
                                Math-Shepherd~\citep{wang2024math}{,}
                                MiPS~\citep{DBLP:conf/emnlp/WangLWLH0S24}{,}
                                OmegaPRM~\citep{DBLP:journals/corr/abs-2406-06592}{,}
                                PQM~\citep{DBLP:journals/corr/abs-2410-11287}{,}
                                AutoPSV~\citep{DBLP:conf/nips/LuD0CDFG24}{,}
                                PAV~\citep{DBLP:journals/corr/abs-2410-08146}{,}
                                Implicit PRM~\citep{DBLP:journals/corr/abs-2412-01981}{,}
                                PRIME~\citep{cui2025process}{,}
                                Qwen-2.5-Math-PRM~\citep{prmlessons}
                                , leaf, text width=34.1em
                            ]
                        ]
                    ]
                ]
                [
                    Usage~(\S\ref{2_3_usage}), node
                    [
                        Data Selection, node
                        [
                            RAFT~\citep{DBLP:journals/tmlr/Dong0GZCPDZS023}{,} 
                            RRHF~\citep{DBLP:conf/nips/YuanYTWHH23}{,} 
                            ReST~\citep{DBLP:journals/corr/abs-2308-08998}{,}
                            Iterative RPO~\citep{DBLP:conf/nips/PangYHCSW24}
                            , leaf, text width=44.6em
                        ]
                    ]
                    [
                        Policy Training, node
                        [   
                            ODIN~\citep{chen2024odin}{,}
                            PCRM~\citep{DBLP:conf/cncl/ZhouWHXZZ24}{,}
                            \citet{DBLP:conf/acl/ParkREF24}{,}
                            CRM~\citep{DBLP:journals/corr/abs-2501-09620}{,}
                            RRM~\citep{liu2025rrm}{,}
                            Bayesian RM~\citep{yang2024bayesian}{,}
                            DPRM~\citep{DBLP:journals/corr/abs-2402-09764}{,}
                            BRME~\citep{yan2024rewardrobustrlhfllms}{,}
                            Fine-Grained RLHF~\citep{wu2024fine}{,}
                            WARM~\citep{ramé2024warmbenefitsweightaveraged}{,}
                            \citet{zhang2024improvingreinforcementlearninghuman}
                            , leaf, text width=44.6em
                        ]
                    ]
                    [
                        Inference, node
                        [
                            ~\citet{DBLP:journals/corr/abs-2310-10080}{,}
                            Tree-PLV~\citep{DBLP:conf/emnlp/He0ZT024}{,}
                            ORPS~\citep{DBLP:journals/corr/abs-2412-15118}{,}
                            ReST-MCTS*~\citep{DBLP:conf/nips/ZhangZHYD024}{,}
                            PAV~\citep{setlur2024rewarding}{,} 
                            STILL-1~\citet{jiang2024technical}{,}
                            RSD~\citet{liao2025reward}{,}
                            , leaf, text width=44.6em
                        ]
                    ]
                ]
                [
                    Application~(\S\ref{4_application}), node
                    [
                        Dialogue, node
                        [
                            Harmless~\citep{bai2022training,glaese2022improving,DBLP:journals/corr/abs-2212-08073,DBLP:conf/iclr/DaiPSJXL0024,DBLP:journals/corr/abs-2412-16339}{,}
                            Professionalism~\citep{zhang2023huatuogpt,yang2024zhongjing}{,}
                            Dialogue Impression~\citep{DBLP:journals/corr/abs-2408-02976,DBLP:journals/corr/abs-2501-12698}
                            , leaf, text width=44.6em
                        ]
                    ]
                    [
                        Reasoning, node
                        [
                            Math~\citep{DBLP:journals/corr/abs-2406-06592,wang2024math,DBLP:conf/iclr/LightmanKBEBLLS24,DBLP:journals/corr/abs-2406-14024,shao2024deepseekmath,zhang2025lessons,zhu2025retrievalaugmentedprocessrewardmodel}{,}
                            Code~\citep{DBLP:journals/corr/abs-2310-10080,DBLP:journals/corr/abs-2406-20060,DBLP:journals/corr/abs-2410-17621,nichols2024performance,DBLP:journals/corr/abs-2409-06957,mcaleese2024llm}
                            , leaf, text width=44.6em
                        ]
                    ]
                    [
                        Retrieve \&\\
                        Recommendation, node
                        [
                            Retrieve~\citep{DBLP:conf/emnlp/0002DW23,xiong2024search, kim2025syntriever,guan2025deepragthinkingretrievalstep,xiong2025raggymoptimizingreasoningsearch}{,}
                            Recommendation~\citep{DBLP:conf/sigir/WangKAJ24,DBLP:journals/corr/abs-2410-05939,DBLP:conf/aaai/0002ZWCZ0HZY24}
                            , leaf, text width=44.6em
                        ]
                    ]
                    [
                        Other \\Applications, node
                        [
                            Text to Audio~\citep{DBLP:conf/icml/CideronGVVKBMUB24, DBLP:conf/ijcai/LiaoHYD0XXLL024,DBLP:journals/corr/abs-2405-14632}{,}
                            Text to Image~\citep{DBLP:journals/corr/abs-2302-12192,DBLP:conf/nips/XuLWTLDTD23,DBLP:journals/corr/abs-2305-16381}{,}
                            Text to Video~\citep{DBLP:conf/nips/WuHWXW24a,DBLP:conf/cvpr/YuanZW0FPZ0A024,wang2025harnesslocalrewardsglobal}{,}
                            Robotic Manipulation~\citep{DBLP:journals/corr/abs-2311-02379,DBLP:conf/iclr/RocamondeMNPL24}{,}
                            Games~\citep{DBLP:journals/corr/abs-2307-12158,choudhury2025processrewardmodelsllm}
                            , leaf, text width=44.6em
                        ]
                    ]
                ]
                [
                    Evaluation~(\S\ref{5_evaluation}), node
                    [
                        Benchmarks, node
                        [
                            RewardBench~\citep{DBLP:journals/corr/abs-2403-13787}{,}
                            RM-Bench~\citep{DBLP:journals/corr/abs-2410-16184}{,}
                            ProcessBench~\citep{DBLP:journals/corr/abs-2412-06559}{,}
                            RMB~\citep{DBLP:journals/corr/abs-2410-09893}{,}
                            VL-RewardBench~\citep{DBLP:journals/corr/abs-2411-17451}{,}
                            PPE~\citep{DBLP:journals/corr/abs-2410-14872}{,}
                            M-RewardBench~\citep{DBLP:journals/corr/abs-2410-15522}{,}
                            MJ-Bench~\citep{DBLP:journals/corr/abs-2407-04842}{,}
                            RAG-RewardBench~\citep{Jin2024RAGRewardBenchBR}{,}
                            PRMBench~\citep{song2025prmbench}{,}
                            Multimodal RewardBench~\citep{yasunaga2025multimodalrewardbenchholisticevaluation}
                            , leaf, text width=44.6em
                        ]
                    ]
                ]
                [
                    Challenges~(\S\ref{6_challenges}), node
                    [
                        Data, node
                        [
                            \citet{DBLP:journals/corr/abs-1909-08593}{,}
                            \citet{DBLP:journals/corr/abs-2211-06519}{,}
                            \citet{DBLP:conf/aaai/BarnettFS023}{,}
                            \citet{DBLP:conf/iclr/BansalDG24}{,}
                            \citet{DBLP:journals/corr/abs-2404-07503}
                            , leaf, text width=44.6em
                        ]
                    ]
                    [
                        Training, node
                        [
                            \citet{DBLP:journals/corr/abs-2009-01325}{,}
                            \citet{DBLP:journals/corr/abs-2209-13085}{,}
                            \citet{gao2023scaling}{,}
                            \citet{weng2024rewardhack}{,}
                            \citet{laidlawcorrelated}{,}
                            \citet{mckee2024honesty}{,}
                            \citet{DBLP:journals/corr/abs-2406-10162}{,}
                            \citet{DBLP:journals/corr/abs-2409-12822}{,}
                            \citet{DBLP:conf/iclr/SharmaTKDABDHJK24}{,}
                            \citet{coste2024rewardmodelensembleshelp}{,}
                            \citet{zhang2024overcomingrewardoveroptimizationadversarial}{,}
                            \citet{miao2025energylossphenomenonrlhf}{,}
                            \citet{liu2025rrm}
                            , leaf, text width=44.6em
                        ]
                    ]
                    [
                        Bias in Evaluation, node
                        [
                            \citet{DBLP:journals/corr/abs-2403-02839}{,}
                            \citet{DBLP:journals/corr/abs-2409-11704}{,}
                            \citet{park2024offsetbias}{,}
                            \citet{DBLP:journals/corr/abs-2502-01534}
                            , leaf, text width=44.6em
                        ]
                    ]
                ]
            ]
        \end{forest}
    }
    \caption{Full taxonomy of Reward Models.}
    \label{fig:taxonomy_full}
\end{figure*}

\subsection{Evaluation Aspects}

According to the benchmarks introduced in (\S\ref{5_evaluation}), the evaluation aspects of RMs can be summarized mainly as follows:

\paragraph{Consistency.} The aim of RMs is to provide preference signals to LLMs, thus consistency is the primary evaluation aspect for RMs. Furthermore, consistency can be divided into: (1) the alignment between RMs and human preferences, the RMs are required to distinguish between chosen and rejected samples~\citep{DBLP:journals/corr/abs-2403-13787,DBLP:journals/corr/abs-2410-16184,DBLP:journals/corr/abs-2410-09893} 
, or identify the correctness of samples directly~\citep{DBLP:journals/corr/abs-2412-06559};(2) the alignment between RMs and policy models, such as style-controlled correlation~\citep{DBLP:journals/corr/abs-2410-16184} and downstream task correlation~\citep{DBLP:journals/corr/abs-2410-14872,DBLP:journals/corr/abs-2411-17451}

\paragraph{Robustness.} On the basis of consistency, RMs should exhibit robustness across the experimental settings and tasks. \citet{DBLP:journals/corr/abs-2410-05584} rewrite the prompts in the RM test dataset to investigate the influence of the prompt semantic bias.
In PRM evaluation, 
\citet{song2025prmbench} requires LLMs to be sensitive to the details of reasoning, including subtle conditions, deception, and multiple solutions.

\paragraph{Safety.}
Similar to the consistency evaluation, \citet{DBLP:journals/corr/abs-2403-13787} and \citet{DBLP:journals/corr/abs-2410-16184} evaluate RM's ability to distinguish between safe and unsafe responses.
\citet{DBLP:journals/corr/abs-2410-09893} conducts trade-off analysis between the goals of helpfulness and harmlessness.

\end{document}